\pdfoutput=1

\documentclass[11pt]{article}

\usepackage[]{ACL2023}

\usepackage{times}
\usepackage{latexsym}

\usepackage[T1]{fontenc}

\usepackage[utf8]{inputenc}

\newcommand{\myparagraph}[1]{\smallskip \noindent{\bf {#1}.}}
\usepackage{microtype}
\usepackage{algorithm} 
\usepackage{amsmath}
\usepackage{times}
\usepackage{latexsym}
\usepackage{extarrows}
\usepackage{hyperref}
\usepackage{url}
\usepackage{booktabs}
\usepackage{enumitem}
\include{bm}
\usepackage{bbding}
\usepackage{graphicx} 

\usepackage{amssymb}
\usepackage{longtable}
\usepackage{setspace}
\usepackage{url}
\usepackage{bm}
\usepackage{tabularx}
\usepackage{xcolor}
\usepackage{multirow}
\usepackage{subfigure}
\usepackage{colortbl}
\usepackage[T1]{fontenc}
\usepackage[utf8]{inputenc}
\usepackage{microtype}

\usepackage{pifont}  
\usepackage{bbding} 
\usepackage{fontawesome} 
\usepackage{inconsolata}

\title{Dual-Gated Fusion with Prefix-Tuning for Multi-Modal Relation Extraction}

\author{
Qian Li$^{1,2}$, Shu Guo$^{3}$, Cheng Ji$^{1,2}$, Xutan Peng$^{4}$, Shiyao Cui$^{5}$, Jianxin Li$^{1,2}$\thanks{\, Corresponding author.}\,, Lihong Wang$^{3}$  \\  
$^{1}$School of Computer Science and Engineering, Beihang University, Beijing, China  \\
$^{2}$Beijing Advanced Innovation Center for Big Data and Brain Computing, Beijing, China \\
$^{3}$National\! Computer\! Network\! Emergency\! Response\! Technical\! Team/Coordination\! Center\! of\! China \\
$^{4}$The University of Sheffield, South Yorkshire, UK \\
$^{5}$Institute of Information Engineering, Chinese Academy of Sciences, Beijing, China\\
\tt \{liqian, jicheng, lijx\}@act.buaa.edu.cn, guoshu@cert.org.cn, \\ 
\tt x.peng@shef.ac.uk, cuishiyao@iie.ac.cn, wlh@isc.org.cn \\
}

\begin{document}
\maketitle
\begin{abstract}
Multi-Modal Relation Extraction (MMRE) aims at identifying the relation between two entities in texts that contain visual clues. Rich visual content is valuable for the MMRE task, but existing works cannot well model finer associations among different modalities, failing to capture the truly helpful visual information and thus limiting relation extraction performance. 
In this paper, we propose a novel MMRE framework to better capture the deeper correlations of text, entity pair, and image/objects, so as to mine more helpful information for the task, termed as DGF-PT.
We first propose a prompt-based autoregressive encoder, which builds the associations of intra-modal and inter-modal features related to the task, respectively by entity-oriented and object-oriented prefixes.
To better integrate helpful visual information, we design a dual-gated fusion module to distinguish the importance of image/objects and further enrich text representations. In addition, a generative decoder is introduced with entity type restriction on relations, better filtering out candidates. Extensive experiments conducted on the benchmark dataset show that our approach achieves excellent performance compared to strong competitors, even in the few-shot situation.
\end{abstract}


\section{Introduction}
\label{sec:intro}
As a fundamental subtask of information extraction, relation extraction (RE) aims to identify the relation between two entities~\cite{DBLP:conf/sigir/CongSC0L022,icassp2}.
Recently, there is a growing trend in multi-modal relation extraction (MMRE), aiming to classify textual relations of two entities as well as introduce the visual contents. It provides additional visual knowledge that incorporates multi-media information to support various cross-modal tasks such as the multi-modal knowledge graph construction~\cite{zhu2022multi, wang2019richpedia} and visual question answering systems ~\cite{DBLP:conf/emnlp/WangMLZLQ022, shih2016look}.

\begin{figure}[t]
    \centering
    \includegraphics[width=\linewidth]{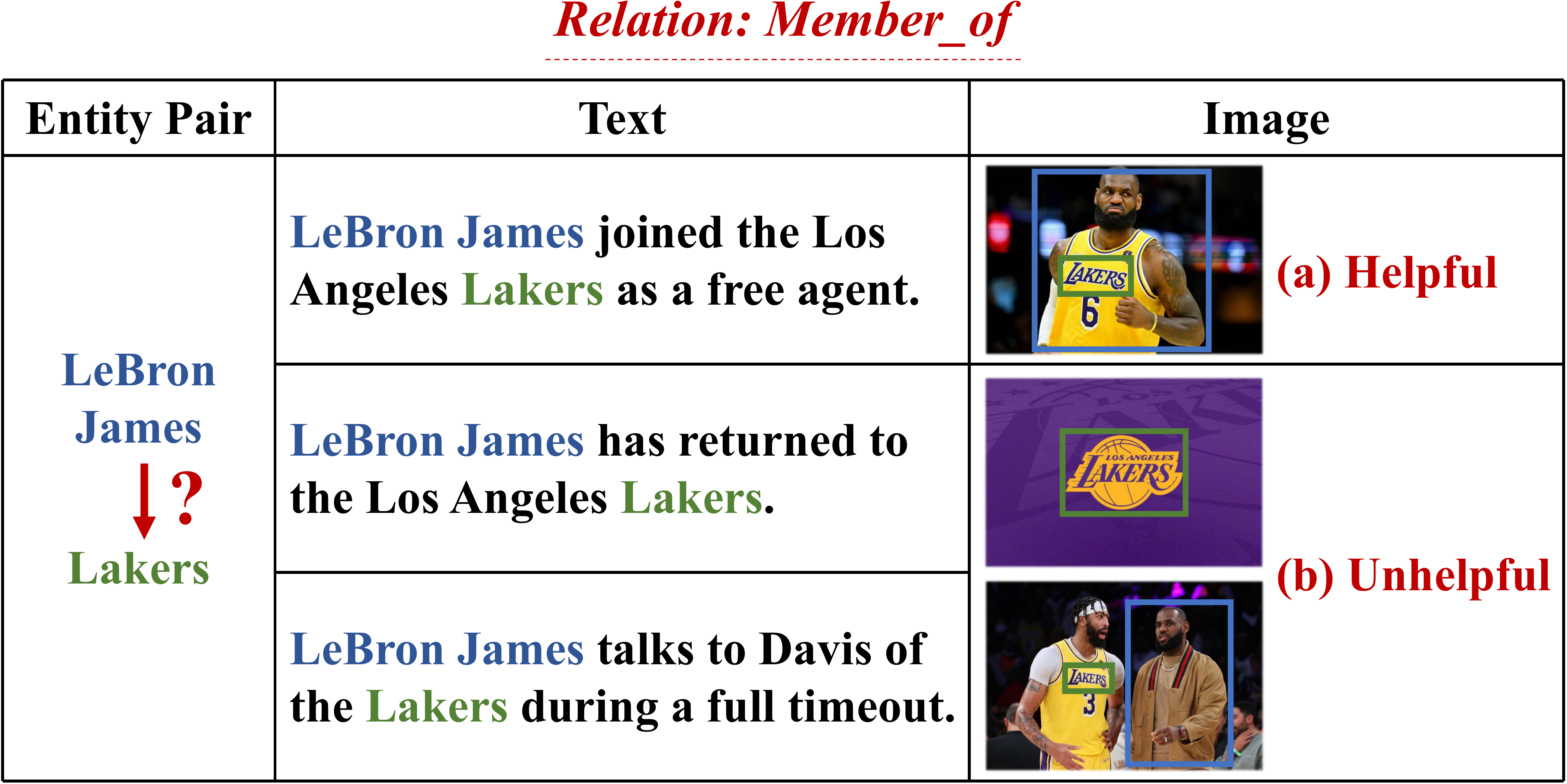}
    \caption{An example of the MMRE task. The task is to predict the relation of given entity pairs for the specific text and image which contains multiple objects.}
    \label{define}
\end{figure}

Existing methods achieved considerable success by leveraging visual information~\cite{DBLP:conf/mm/ZhengFFCL021, DBLP:journals/corr/abs-2211-05528, DBLP:conf/naacl/ChenZLYDTHSC22} since the visual contents provide valuable pieces of evidence to supplement the missing semantics for MMRE. 
Previous work~\cite{DBLP:conf/mm/ZhengFFCL021} introduced the visual relations of objects in the image to enrich text embedding via an attention-based mechanism.
Next, HVPNet~\cite{DBLP:conf/naacl/ChenZLYDTHSC22} used an object-level prefix and a multi-scale visual fusion mechanism to guide the text representation learning. 
Nevertheless, these methods primarily focus on the relations between objects and text and ignore the finer associations (entity pair, text, and image/objects). Furthermore, they usually suffered from the failure of identifying truly helpful parts/objects of the image to the corresponding entity pair on account of introducing all the objects. This may cause severe performance degradation of downstream tasks.


For multi-modal relation extraction, not all images or their objects are helpful for prediction. As illustrated in Figure \ref{define}, given three different inputs with the same relation \textit{Member\_of} and entity pair, each of the inputs contains a text, an image, and an entity pair. 
There are two situations: (a) The image is helpful for relation extraction. For entity pair \textit{LeBron James} and \textit{Lakers}, in the image \textit{LeBron James} wears the Lakers jersey revealing the implied relationship between the two entities. 
Therefore, we can improve relation extraction by considering the entity-pair relationships in visual information.
(b) The image is unhelpful for the entity pair \textit{LeBron James} and \textit{Lakers} since it only contains \textit{Lakers} object, rather than the association information for entity pairs. Furthermore, the image can provide an incorrect extraction signal, for example, the third image in Figure \ref{define} shows that the relation between \textit{LeBron James} and \textit{Lakers} is more likely to be misjudged as \textit{Coach\_of} or \textit{Owner\_of}. Unhelpful visual content is prone to providing misleading information when predicting the relation.
In general, it is necessary to identify the truly helpful visual information to filter the useless and misleading ones, but it is still under-explored.


To overcome the above challenges, we propose a novel 
MMRE framework DGF-PT to better incorporate finer granularity in the relations
and avoiding unhelpful images misleading the model\footnote{The source code is available at \url{https://github.com/xiaoqian19940510/DGF-PT}.}.
Specifically, we propose a prompt-based autoregressive encoder containing two types of prefix-tuning to integrate deeper associations.
It makes the model focus on associations of intra-modal (between entity pair and text) by entity-oriented prefix and inter-modal (between objects and text) by the object-oriented prefix.
In order to distinguish the importance of image/objects, we design a dual-gated fusion module to address the unhelpful visual data by utilizing interaction information via local and global visual gates.
Later, we design a generative decoder to leverage the implicit associations and restrict candidate relations by introducing entity type.
We further design joint objective to allow the distribution of representations pre and post-fusion to be consistent while enhancing the model to identify each sample in the latent space.
Experimental results show that our approach achieves excellent performance in the benchmark dataset.
Our contributions can be summarized as follows. 
\begin{itemize}[leftmargin=*]
    \item We technically design a novel MMRE Framework to build deeper correlations among entity pair, text, and image/objects and distinguish helpful visual information.
    \item We propose a prompt-based autoregressive encoder with two types of prefixes to enforce the intra-modal and inter-modal association. We design dual-gated fusion with a local object-importance gate and a global image-relevance gate to integrate helpful visual information. 
    \item Experimental results indicate that the framework achieves state-of-the-art performance on the public multi-modal relation extraction dataset, even in the few-shot situation.
    
\end{itemize}

\begin{figure*}[t]
    \centering
    \includegraphics[width=\linewidth]{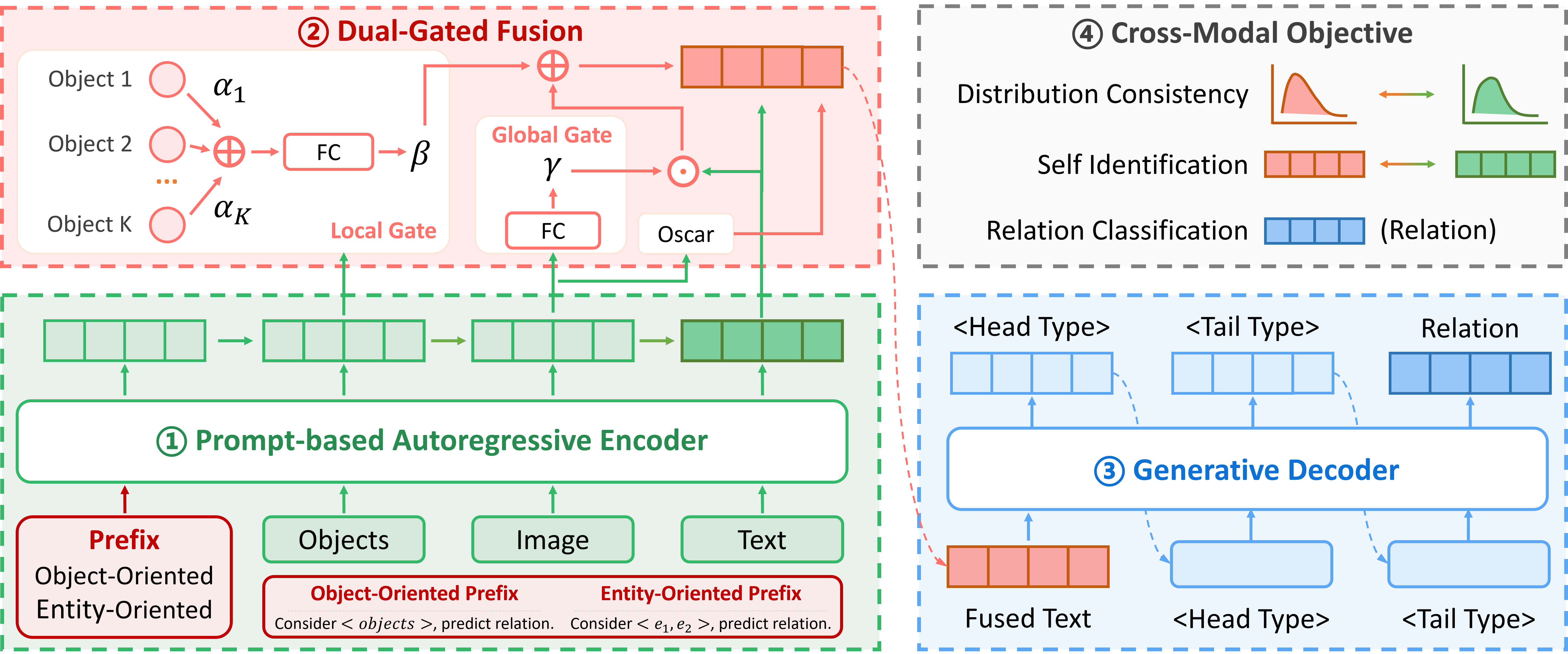}
    \caption{The DGF-PT framework for multi-modal relation extraction.}
    \label{Framework}
\end{figure*}

\section{Related Work}

Multi-modal relation extraction (MMRE) task, a subtask of multi-modal information extraction in NLP~\cite{DBLP:conf/aaai/0006W0SW21, DBLP:conf/pkdd/Cong0LCT020, DBLP:conf/acl/SuiT00020,DBLP:conf/acl/0001LDXLHSW22}, aims to identify textual relations between two entities in a sentence by introducing visual content~\cite{DBLP:conf/mm/ZhengFFCL021,DBLP:conf/icmcs/ZhengWFF021,DBLP:conf/naacl/ChenZLYDTHSC22}, which compensates for insufficient semantics and helps to extract the relations.

Recently, there are several works~\cite{DBLP:conf/mm/ZhengFFCL021,DBLP:conf/icmcs/ZhengWFF021,DBLP:conf/naacl/ChenZLYDTHSC22} beginning to focus on the multi-modal relation extraction technology.
As the first work, MNRE~\cite{DBLP:conf/icmcs/ZhengWFF021} developed a multi-modal relation extraction baseline model. It demonstrates that introducing multi-modal information supplements the missing semantics and improves relation extraction performance in social media texts.
Later, \citet{DBLP:conf/mm/ZhengFFCL021} proposed a multi-modal neural network containing a scene graph modeling the visual relations of objects and aligning the relations between objects and text via similarity and attention mechanism. 
HVPNet~\cite{DBLP:conf/naacl/ChenZLYDTHSC22} designs a visual prefix-guided fusion for introducing object-level visual information and further utilizes hierarchical multi-scaled visual features.
Moreover, they introduce the information of all objects and thus cannot distinguish the truly helpful visual information, making it impractical to personalize the use of the image and further damaging their performance.


In the multi-modal relation extraction task, we note that images are naturally helpful information in the problem of multi-modal relation extraction. However, the potential for a differentiated use of image information in this task is under-explored. 
In this paper, we focus on the finer (intra-modal and inter-modal) association and manage to integrate truly useful visual information, promoting the exploitation of limited images. This enables bridging the gap through the transfer multi-modal relation extraction task into MLM pre-train mechanism \cite{DBLP:conf/naacl/DevlinCLT19, DBLP:journals/corr/abs-2107-13586}. 

\section{Problem Formulation}

We provide the definition of MMRE.
For a given sentence text $T=\{w_1, w_2, \ldots, w_L\}$ with $L$ words as well as the image $I$ related to the sentence, and an entity pair $(e_1, e_2)$, an MMRE model takes $(e_1, e_2, T, I)$ as input and calculates a confidence score $p(r_i|e_1, e_2, T, I)$ for each relation $r_i \in R$ to estimate whether $T$ and $I$ can reflect the relation $r_i$ between $e_1$ and $e_2$. $R=\{r_1, \ldots, r_C, \text{None}\}$ is a pre-defined relation set, where ``None'' means that no relation is maintained for the mentions.

\section{Framework}

This section introduces our proposed DGF-PT framework, as shown in Figure~\ref{Framework}. 
We first design the prompt-based autoregressive encoder to acquire the fine-grained representations (entity pair, object, and text), which contains two types of prefixes for integrating helpful information and characterizing intra-modal and inter-modal interactions.
To avoid unhelpful visual information misleading the model, we design the dual-gated fusion module to distinguish the importance of image/objects via local and global gates. It also integrates the semantics of image transferred by Oscar~\cite{DBLP:conf/eccv/Li0LZHZWH0WCG20}. The Fusion module outputs an enhanced representation.
Later, the generative decoder is proposed for relation prediction, leveraging the implicit associations and restricting candidate relations by introducing entity types.
Finally, we design the joint objective including distribution-consistency constraint, self-identification enhancement, and relation classification for model optimization.


\subsection{Prompt-based Autoregressive Encoder}
In order to acquire the finer granularity in the associations (entity pair, objects, and text), we propose a prompt-based autoregressive encoder. After initialization, two specific prefix-tuning strategies are implemented to guide the encoder to attend to task-relevant inter-/intra-modal associations. Subsequently, prefixes, objects, image, and text are progressively fed into an autoregressive encoder stage by stage to obtain fine-grained representations for use in subsequent fusion module (Section~\ref{Alignment}).

\subsubsection{Initialization}
Given text $T$, the word embeddings $\mathbf{w} \in \mathbb{R}^{1 \times N}$ are obtained through the GPT-2 model ~\cite{radford2019language} and then fed into a fully-connection layer, where $N$ is the word dimension. The initial text representation is $\mathbf{T}=\left[\mathbf{w}_1 ; \mathbf{w}_2 ; \ldots ; \mathbf{w}_L\right] \in \mathbb{R}^{L \times N}$.

Given an image $I$, the global image feature $\mathbf{I} \in \mathbb{R}^{M \times N}$ is obtained by VGG16~\cite{DBLP:journals/corr/SimonyanZ14a} with a fully-connection layer, transferring the feature into $M$-block $N$-dimensional vectors. We then extract object features using Faster R-CNN~\cite{DBLP:conf/nips/RenHGS15} and select top-$K$ objects using the ROI classification score. Each object feature is obtained by the average grouping of ROI regions. The initial object representation is $\mathbf{O}=\left[\mathbf{o}_1 ; \mathbf{o}_2 ; \ldots ; \mathbf{o}_K\right] \in \mathbb{R}^{K \times N}$.

\subsubsection{Object \& Entity Oriented Prefixes}

Utilizing the advantages of prefix-tuning, the pre-trained encoder (e.g., GPT-2 as our encoder) can be guided to learn task-specific features for fast adaptation to the MMRE task~\cite{DBLP:journals/corr/abs-2107-13586}. However, the design of appropriate prefixes for finer associations learning in the MMRE task remains an open research question, and the direct use of prefixes from other tasks is not reasonable. Therefore, we construct two types of prefixes, an object-oriented prefix for inter-modal relevance (objects \& text) and an entity-oriented prefix for intra-modal correlations (entity pair \& text), encouraging the encoder to leverage text as a medium to strengthen multi-granular associations to acquire enhanced semantic representations.

\myparagraph{Object-Oriented Prefix} 
Given that objects related to entities are indeed useful information for the MMRE task, we propose an object-oriented prefix, termed as $\mathcal{P}_o(\cdot)$, which provides guidance information of inter-modal relevance to the encoder.
For the input text $T$, we define the following pattern ``Consider $\langle objects\rangle$, predict relation.'', where $\langle objects\rangle$ means the objects relevant to the entity pair of $T$ which is different for each input.
It emphasizes specific key textual contents and introduces the visual features of relevant objects.

\myparagraph{Entity-Oriented Prefix} 
Due to the visual information may be incomplete or misleading, we argue that only an object-oriented prefix is insufficient to capture classification information. Thus, we propose an entity-oriented prefix, termed as $\mathcal{P}_e(\cdot)$,  to capture intra-modal association to adapt the task. We define the following pattern ``Consider $\langle {e_{1},e_{2}}\rangle$, predict relation.'', where $\langle {e_{1},e_{2}}\rangle$ is the entity pair to predict the relation.

\subsubsection{Multi-Stage Autoregressive Encoder}
Prompt-based learning keeps the parameters of the whole PLM frozen and prepends the prefixes before the task inputs \cite{DBLP:journals/corr/abs-2107-13586}.
The bidirectional encoder (e.g., BERT) cannot effectively integrate the proposed dual-gated fusion module (Section \ref{Alignment}) in model testing.
Therefore, we deploy a unidirectional encoder (e.g., GPT and GPT-2) and design multiple stages to integrate multi-granular textual and visual knowledge, where the prefixes, objects, image, and text are fed stage by stage. 

\myparagraph{First stage ($S_1$)}
The input of the first stage $S_1$ contains the prefixes and objects to learn the relevance from local granularity and obtain the representations of objects.
To introduce task-related prefix knowledge, the two types of trainable prefixes $\mathcal{P}_o(\cdot)$ and $\mathcal{P}_e(\cdot)$ are prepended before the input sequence as the prefix tokens, obtained through the GPT-2 vocabulary. In $S_1$, the encoder learns the representations of objects and updates the prefix tokens of each model layer:
\begin{equation}
\begin{aligned}    
&\mathcal{P}^*_o(\mathbf{o}_{e_1},\mathbf{o}_{e_2}), \mathcal{P}^*_e(\mathbf{T}[e_1],\mathbf{T}[e_2]), \mathbf{h}_o
=\mathcal{S}_1(\cdot)\\
&=\!\operatorname{Encoder}\!\left(\mathcal{P}_o(\mathbf{o}_{e_1},\!\mathbf{o}_{e_2}), \mathcal{P}_e(\mathbf{T}[e_1],\!\mathbf{T}[e_2]), \!\mathbf{O} \right),
\end{aligned}
\end{equation}
where $\mathcal{P}^*_o(\cdot)$ and $\mathcal{P}^*_e(\cdot)$ are updated prefixes after $S_1$, $\mathbf{h}_o$ is the representations of objects, and $\mathbf{o}_{e_1}$ and $\mathbf{o}_{e_1}$ are the initial embeddings of entities $e_1$ and $e_2$.
After the $S_1$ stage, the object information is introduced into the prefix embedding.

\myparagraph{Second stage ($S_2$)}
The inputs of the second stage $S_2$ are the outputs of the first stage $\mathcal{S}_1$ (including the updated prefixes and the representations of objects) and the image feature $\mathbf{I}$ to get the representations of images $\mathbf{h}_i$.
We hope the model can learn to capture the inter-modal relevance from global granularity, which is useful information for relation extraction that may improve performance. Thus, we introduce the image information in $S_2$. 
The $S_2$ embedding is therefore updated as:
\begin{equation}
\begin{aligned}    
\mathbf{h}_i=\mathcal{S}_2(\cdot)=\operatorname{Encoder}\left(\mathcal{S}_1(\cdot), \mathbf{I} \right).
\end{aligned}
\end{equation}

\myparagraph{Third stage ($S_3$)}
To learn text representation $\mathbf{h}_t$, the third stage inputs $\mathcal{S}_2$ and text $T$  using interactive objects and images. 
\begin{equation}
\begin{aligned}    
\mathbf{h}_t=\mathcal{S}_3(\cdot)=\operatorname{Encoder}\left(\mathcal{S}_2(\cdot), \mathbf{T} \right).
\end{aligned}
\end{equation}





\subsection{Dual-Gated Fusion}
\label{Alignment}
Unhelpful/Task-irrelevant information in image is often ignored by simply utilizing all objects for aggregation. 
To solve this, we propose dual-gated fusion to effectively integrate helpful visual information while filtering out misleading information. This module utilizes local and global gates to distinguish the importance and relevance of image/objects, and filters out task-irrelevant parts. By integrating semantic information of the image, a final fused representation containing associations among image, object, and text is obtained.

Specifically, the local object-importance gate vector $\bm{\beta}$ by the local object features and the global image-relevance gate vector $\bm{\gamma}$ by the global image features are calculated as:
\begin{align}
\alpha_k &= \frac{\cos \left(\mathbf{h}_t, \mathbf{h}_o[k]\right)}{\sum_{j=1}^K \cos \left(\mathbf{h}_t, \mathbf{h}_o[j]\right)}, \\
    \bm{\beta} &= \operatorname{FC_\beta}(\mathbf{\overline{h}}_o)=\operatorname{FC_\beta}\left(\sum_{k=1}^K \alpha_k \mathbf{h}_o[k]\right), \\
    \bm{\gamma} &= \operatorname{tanh}\left( \operatorname{FC_\gamma}\left( \mathbf{h}_i \right)\right),
\end{align}
where $\operatorname{FC}$ is fully-connected layer and $\mathbf{h}_o[k]$ is the $k$-th object in the object set $O$.
$\mathbf{\overline{h}}_o$ calculates attention between the selected top-$K$ objects of divergent modalities.
Subsequently, the textual characteristic of fusion $\tilde{\mathbf{h}}_t$ is calculated by
\begin{equation}
\tilde{\mathbf{h}}_t=\operatorname{MLP}\left(\mathbf{h}_t \odot \bm{\gamma}+\bm{\beta}\right)+\mathbf{h}_t,
\end{equation}
where $\operatorname{MLP}$ is multilayer perceptron and $\odot$ means hadamard product.


In order to further integrate the semantics of visual information, we use Oscar~\cite{DBLP:conf/eccv/Li0LZHZWH0WCG20} to transfer $\mathbf{h}_i$ into a text description $\tilde{\mathbf{h}}_{i2t}$ for each image, using objects as anchor points to align visual and textual features in a common space. It learns multi-modal alignment information of entities from a semantic perspective.
The detail is given in Appendix \ref{Oscar}. 

While local representations can capture valuable clues, global features provide condensed contextual and high-level semantic information. Given this insight, we leverage the global information from one modality to regulate the local fragment of another modality, enabling the entity to contain semantic information and filter out irrelevant visual information. The final fused representation is:
\begin{equation}
\tilde{\mathbf{h}}_t= \tilde{\mathbf{h}}_t +\tilde{\mathbf{h}}_{i2t} \odot \delta,
\end{equation}
where $\delta$ is trade-off factor between text embedding $\tilde{\mathbf{h}}_t$ and the inter-modal text representation $\tilde{\mathbf{h}}_{i2t}$. 

\subsection{Generative Decoder}

To leverage the implicit associations and restrict candidate relations by introducing entity type, we design a generative decoder.

The type of entity pair is helpful for relation classification. For example, the relation of entity type \textit{Person} and \textit{Organization} must not be \textit{born} and \textit{friend}, but maybe \textit{CEO} and \textit{staff}.
Thus, we introduce head type $\mathbf{T}_{e_1}^{t}$ and tail type $\mathbf{T}_{e_2}^{t}$ one by one
to leverage the implicit associations and restrict candidate relations.

To maintain the consistency of the relation extraction task with the MLM pre-trained model, we use the generative decoder to predict the relation.  
The prediction of the generative decoder is:
\begin{equation}
\mathbf{h}_{e_1}^{t},\mathbf{h}_{e_2}^{t}, \mathbf{r}=\operatorname{Decoder}\left(\mathcal{S}_3(\cdot),\mathbf{T}_{e_1}^{t},\mathbf{T}_{e_2}^{t} \right),
\end{equation}
where $\mathbf{h}_{e_1}^{t},\mathbf{h}_{e_2}^{t}$ are the representation of types, and $\mathbf{r}$ is the representation of relation.

\subsection{Joint Objective}

In order to address distribution consistency within the dual-gated fusion module, we introduce the distribution-consistency constraint loss, which is applied on a single-sample basis. Additionally, to meet the need for inter-sample identification, we propose self-identification enhancement loss. The overall joint objective is then formed by combining the relation classification loss with the aforementioned constraints.

\paragraph{Distribution-Consistency Constraint.}
In order to ensure the dual-gated fusion module effectively integrates helpful visual features while avoiding the introduction of task-irrelevant information, we introduce distribution-consistency constraint to measure and optimize the change in representation distribution pre and post-fusion. 
Thus, we propose to use KL divergence to measure the distance between the probability distribution of $\tilde{\mathbf{h}}_t$ and $\mathbf{h}_t$, which is equal to calculating the cross-entropy loss over the two distributions:
\begin{align}
\label{eq3}
    \mathcal{L}_{d}(\theta) & =\mathrm{KL}\left(p_\theta(\boldsymbol{r} | \tilde{\mathbf{h}}_t) \| p_\theta(\boldsymbol{r} | \mathbf{h}_t)\right) \\
    &=\!\!\sum_{\boldsymbol{r} \in R} p_\theta(\boldsymbol{r} | \tilde{\mathbf{h}}_t) \log p_\theta(\boldsymbol{r} | \mathbf{h}_t).
\end{align}

\begin{table*}[t]
\centering
\renewcommand\arraystretch{1.15}
\resizebox{\linewidth}{!}{
\begin{tabular}{ll|cccc}
\toprule
  \textbf{Model Type} &\textbf{Model Name}  {  }{  } {  }{  } {  }{  } {  }{  }  {  }{  } {  }{  } {  }{  } {  }{  }  {  }{  } {  }{  } {  }{  } {  }{  }{  }{  } {  }{  }& {  }{  } {  }{  }\textbf{Acc. (\%)}  {  }{  } {  }{  }&  {  }{  } {  }{  }\textbf{Prec. (\%)}   {  }{  } {  }{  }&  {  }{  } {  }{  }\textbf{Recall (\%)}  {  }{  } {  }{  } & \textbf{F1 (\%)} \\
\midrule
& \textbf{Glove+CNN~\cite{DBLP:conf/coling/ZengLLZZ14} }   & 70.32  & 57.81  & 46.25  & 51.39   \\  
\textbf{Text-based RE Models} & \textbf{PCNN~\cite{DBLP:conf/emnlp/ZengLC015} }  & 72.67  & 62.85  &49.69 & 55.49      \\  
& \textbf{MTB~\cite{DBLP:conf/acl/SoaresFLK19}}   & 72.73  & 64.46  & 57.81  & 60.96   \\   \midrule
& \textbf{BERT+SG~\cite{DBLP:conf/naacl/DevlinCLT19}  }   &  74.09 & 62.95 &  62.65 & 62.80   \\
& \textbf{BERT+SG+Att.~\cite{DBLP:conf/naacl/DevlinCLT19} }   &  74.59 & 60.97 & 66.56 & 63.64   \\ 
\textbf{MMRE Models}& \textbf{VisualBERT~\cite{DBLP:journals/corr/abs-1908-03557} } {  }{  } {  }{  } {  }{  } {  }{  } & -  &  57.15 & 59.45 & 58.30  \\ 
& \textbf{MEGA~\cite{DBLP:conf/mm/ZhengFFCL021} } {  }{  } {  }{  } {  }{  } {  }{  } & \underline{76.15}  &  64.51 & 68.44 & 66.41  \\ 
    
 &   \textbf{HVPNet~\cite{DBLP:conf/naacl/ChenZLYDTHSC22} } {  }{  } {  }{  } {  }{  } {  }{  } & -  &  \underline{83.64} & \underline{80.78} & \underline{81.85}  \\ \midrule

& \textbf{DGF-PT (BERT Encoder) }   & 79.82 ( $\uparrow$ 3.67 )  &  79.72 ( $\uparrow$ -3.92 ) & 78.63 ( $\uparrow$ -2.15 ) & 79.24 ( $\uparrow$ -2.61 )   \\
\textbf{Ours}&  \textbf{DGF-PT (GPT Encoder) }   & 82.03 ( $\uparrow$ 5.88 )  &  81.23 ( $\uparrow$ -2.41 ) & 82.48 ( $\uparrow$ 1.70 ) & 82.09 ( $\uparrow$ 0.24 )    \\
&  \textbf{DGF-PT (GPT-2 Encoder) }   & \textbf{84.25 ( $\uparrow$ 8.10 )}  &  \textbf{84.35 ( $\uparrow$ 0.71 )} & \textbf{83.83 ( $\uparrow$ 3.05 )} & \textbf{84.47 ( $\uparrow$ 2.62 )}   \\
\bottomrule
\end{tabular}
}
\caption{Main experiments. The best results are highlighted in bold, ``–'' means results are not available, and the underlined values are the second-best result. ``$\uparrow$'' means the increase compared to the underlined values. 
}
\label{main}
\end{table*}

\paragraph{Self-Identification Enhancement.}
The MMRE task requires the model to have the ability to correctly classify relations from individual samples. However, relation labels are unevenly distributed or lacking in the real world. Therefore, further enhancement is needed. We design a negative-sampling-based self-supervised loss function to enhance the model. Moreover, the dual-gated fusion module is treated as the augmentation function leveraging the modality information. Specifically, textual representation $\mathbf{h}_t$ and fused representation $\tilde{\mathbf{h}}_{t}$ are the mutually positive samples:
\begin{equation}
\mathcal{L}_s  \!\!=\!\! \left[s(x,\tilde{x})\!-\!s(x_n,\tilde{x}) \right]_+ \!\!\!+\! \left[s(x,\tilde{x})\!-\!s(x,\tilde{x}_n) \right]_+,
\end{equation}
where $\{x,\tilde{x}\}$ are $\{\mathbf{h}_t,\tilde{\mathbf{h}}_t\}$, $[a]_{+}= \max(a,0)$, and $s(\cdot,\cdot)$ is the cosine similarity. $x_n$ and $\tilde{x}_n$ are the hardest negatives of $\mathbf{h}_{t}$ and $\tilde{\mathbf{h}}_t$ in a mini-batch based on a similarity-based measurement.

\paragraph{Relation Classification.}
The loss for relation classification by the negative log-likelihood function is as follows:  
\begin{equation}
\mathcal{L}_{\text {c}}=-\log p(\mathbf{r} | \mathbf{h}_t, \mathbf{h}_i, \mathbf{h}_t[e_1], \mathbf{h}_t[e_2]),
\end{equation}
where $\mathbf{r}$ is the relation between the head entity $e_1$ and the tail entity $e_2$ for $\mathbf{h}_x$, and $\mathbf{h}_t[e_1], \mathbf{h}_t[e_2]$ are the representations of the two entities. Finally, the overall loss function of our model is as follows.
\begin{equation}
\mathcal{L}=\lambda_{d} \mathcal{L}_d+\lambda_s \mathcal{L}_s+\lambda_c \mathcal{L}_c,
\end{equation}
where $\lambda_{d}$, $\lambda_{s}$, and $\lambda_{c}$ are trade-off parameters. We optimize all training inputs in a mini-batch strategy.

\section{Experiment}

\subsection{Dataset and Evaluation Metric}
We conduct experiments in a multi-modal relation extraction dataset MNRE~\cite{DBLP:conf/icmcs/ZhengWFF021}, crawling data from Twitter\footnote{https://archive.org/details/twitterstream}. The MNRE includes $15,484$ samples and $9,201$ images. It contains $23$ relation categories. As previous work~\cite{DBLP:conf/mm/ZhengFFCL021} recommended, the MNRE dataset is divided into $12,247$ training samples, $1,624$ development samples, and $1,614$ testing samples. We report the official Accuracy (Acc.), Precision (Prec.), Recall, and F1 metrics for relation evaluation.

\subsection{Comparision Methods}
We compare our method with three text-based RE models and five MMRE models.

\myparagraph{Text-based RE Models} We first consider a group of representative text-based RE models, which do not introduce image information, for modeling the connection of words in the sentence: (1) Glove+CNN~\cite{DBLP:conf/coling/ZengLLZZ14} is a CNN-based model with additional position embeddings to utilize the position association. (2) PCNN~\cite{DBLP:conf/emnlp/ZengLC015} is a RE method utilizing external knowledge graph with a distant supervision manner to build connection by the graph. (3) Matching the Blanks (MTB)~\cite{DBLP:conf/acl/SoaresFLK19} is a BERT-based RE model to learn context correlation. 

\myparagraph{MMRE Models} We further consider another group of previous approaches for MMRE to integrate visual information: (4) \textbf{BERT+SG}~\cite{DBLP:conf/naacl/DevlinCLT19} concatenates BERT representations with visual content, which is obtained by the pre-trained scene graph tool~\cite{DBLP:conf/cvpr/TangNHSZ20} to learn the connection between text and the object of the image.
(5) \textbf{BERT+SG+Att} adopts an attention mechanism to compute the relevance between the textual and visual features. (6) \textbf{VisualBERT}~\cite{DBLP:journals/corr/abs-1908-03557} is a single-stream encoder, learning cross-modal correlation in a model. (7) \textbf{MEGA}~\cite{DBLP:conf/mm/ZhengFFCL021} considers the relevance from the structure of objects in the image and semantics of text perspectives with graph alignment. (8)   \textbf{HVPNet}~\cite{DBLP:conf/naacl/ChenZLYDTHSC22} introduces an object-level prefix with a dynamic gated aggregation strategy to enhance the correlation between all objects and text.


In contrast to these methods, our approach incorporates the correlation between entity pairs, text, and visual information, and effectively identifies useful visual information.

\subsection{Implementation Details}
For all baselines, we adopt the best hyper-parameters and copy results reported in the literature~\cite{DBLP:conf/mm/ZhengFFCL021, DBLP:conf/icmcs/ZhengWFF021, DBLP:conf/naacl/ChenZLYDTHSC22}.

We used PyTorch\footnote{https://pytorch.org/} as a deep learning framework to develop the MMRE. 
The BERT and GPT-2\footnote{https://github.com/huggingface} are for text initialization and the dimension is set at $768$. 
The VGG version is VGG16\footnote{https://github.com/machrisaa/tensorflow-vgg}. We use Faster R-CNN~\cite{DBLP:conf/nips/RenHGS15} for image initialization and set the dimension of visual objects features at $4096$. 
For hyper-parameters, the best coefficients $ \lambda_{d}, \lambda_{s}, \lambda_{c}$ are $2$, $2$ and $3$. The best $\delta$ is 0.4.
See Appendix~\ref{Settings} for more details on model training. 

\subsection{Main Results}
To verify the effectiveness of our model, we report the overall average results in Table~\ref{main}. 

From the table, we can observe that: 1) Our model outperforms text-based RE models in terms of four evaluation metrics, indicating the beneficial impact of visual information on relation extraction and the necessity of its integration.
2) Compared to MMRE baselines, our model achieves the best results. Specifically, our model improves at least 2.62\% in F1 and 8.10\% in Acc., respectively. 
These results indicate that our method for incorporating and utilizing visual information is superior and effective.
3) Compared to different encoders (e.g., BERT and GPT), the GPT and GPT-2 achieve better results. It demonstrates that the generative encoder can integrate effective visual features more effectively, which is more suitable for the task. For the generative model, the performance is sensitive to the order of input. Thus, we discuss the effect of the order of text, image, and objects in Appendix~\ref{Order}.

\subsection{Discussion for Model Variants}

For a further detailed evaluation of the components of our framework, we performed ablation experiments and reported the results in Table~\ref{Ablation}. E-P means entity-oriented prefix and O-P means object-oriented prefix. ``$\downarrow$'' means the average decrease of all four metrics compared to our model. 

\myparagraph{Discussions for core module} To investigate the effectiveness of each module, we performed variant experiments, showcasing the results in Table~\ref{Ablation}. From the table, we can observe that: 1) the impact of the prefixes tends to be more significant. We believe the reason is that the multiple prompts characterize modality interactions, helping for providing more visual clues. 2) By removing each module, respectively, the performance basically decreased. Compared to joint objective modules, the dual-gated fusion is significantly affected. It demonstrates the effectiveness of knowledge fusion introducing useful visual content and addressing noise visual data.
All observations demonstrate the effectiveness of each component in our model.

\begin{table}[t]

\centering
\renewcommand\arraystretch{1.35}
\resizebox{\linewidth}{!}{
\begin{tabular}{l|ccccc}
\toprule
 \textbf{Variants} & \textbf{Acc.}  &  \textbf{Prec.}   &  \textbf{Recall}   &  \textbf{F1} & \textbf{$\mathbf{\triangle}$ Avg} \\
 \midrule
 \textbf{$\text{  DGF-PT (Ours)}$}  & \textbf{85.25}  & \textbf{84.35} & \textbf{83.83} & \textbf{84.47}  & - \\ \midrule
 {  }{  } \textbf{  w/o All Prefixes}   & 83.32  & 82.93  & 83.42   & 82.35 &$\downarrow$ 1.47  \\  
  {  }{  } \textbf{  w/o E-P}  & 84.62  & 83.31  &82.83  & 82.63  & $\downarrow$ 1.13  \\ 
   {  }{  } \textbf{ w/o O-P}  & 84.05  & 83.20  &83.29  & 83.39  & $\downarrow$ 0.99  \\  
 {  }{  } \textbf{  w/o dual-gated fusion}   & 84.24  & 83.56  & 82.74  & 83.26 & $\downarrow$ 1.03  \\ 
  {  }{  } \textbf{  w/o joint objective}    & 84.49 & 83.90 & 83.26  & 84.05  & $\downarrow$ 0.55   \\ 
\midrule

 {  }{  } \textbf{ repl. All Prefixes in $S_2$}   & 83.29  & 82.40  & 83.17   & 82.24  & $\downarrow$ 1.70  \\  
  {  }{  } \textbf{ repl. All Prefixes in $S_3$}  & 84.10  & 83.24  &82.02  & 82.29  & $\downarrow$ 1.56   \\ 
 {  }{  } \textbf{ repl. E-P in $S_2$}   & 84.28  & 83.20  & 82.41  & 83.93  & $\downarrow$ 1.02 \\ 
 {  }{  } \textbf{ repl. E-P in $S_3$}   & 84.37  & 83.82  &83.03  & 83.41 &  $\downarrow$ 0.82  \\  
  {  }{  } \textbf{ repl. O-P in $S_2$}    & 84.12 & 83.71 & 83.15  & 83.84  & $\downarrow$ 0.77  \\  
 {  }{  } \textbf{ repl. O-P in $S_3$}   &  84.03 & 83.65 & 82.20 &  83.73  & $\downarrow$ 1.07  \\ 

  {  }{  } \textbf{ repl. E-P in $S_2$ \& O-P in $S_3$}    & 84.09 & 83.43 & 82.81  & 84.15 & $\downarrow$ 0.86   \\
  {  }{  } \textbf{ repl. E-P in $S_3$ \& O-P in $S_2$}    & 84.76 & 84.24 & 83.38  & 84.20 &  $\downarrow$ 0.33  \\  
\bottomrule
\end{tabular}
}
\caption{Variant experiments on different orders to introduce the two prefixes. ``w/o'' means removing the corresponding module from the complete model. ``repl.'' means replacing the stage of introducing prefix. }
\label{Ablation}
\end{table}


\myparagraph{Discussions for the stage of prefix}
We explore the effects by introducing the prefixes at a different stage in the encoder, as shown in Table~\ref{Ablation}. From the table, we can observe that: 
1) Compared to feed all Prefixes in the $S_2, S_3$ stage, the $S_1$ stage is more effective. It demonstrates that early introducing prefixes may integrate more helpful visual classification information. 2) When the O-P is fixed, and the E-P is fed into the $S_3$ stage, the performance of our model is the best compared to that introduced in $S_2$ stages. It demonstrates that the E-P is set nearly to the text features helpful to introduce intra-modal association. 3) When we fix the E-P, and the O-P is introduced in the $S_2$ stage, which is fed nearly to objects, the performance achieves better than $S_3$. It demonstrates that the O-P is nearly to object features, which can capture more useful local information for relation classification. 4) When we change the stage of the two prefixes, the performance achieves better as the E-P in $S_3$ and the O-P in $S_2$. 
All observations demonstrate that ``E-P in $S_1$ \& O-P in $S_1$'' is the best schema to introduce intra-modal association and inter-modal relevance.

\begin{figure}[t]
    \centering
    \includegraphics[width=\linewidth]{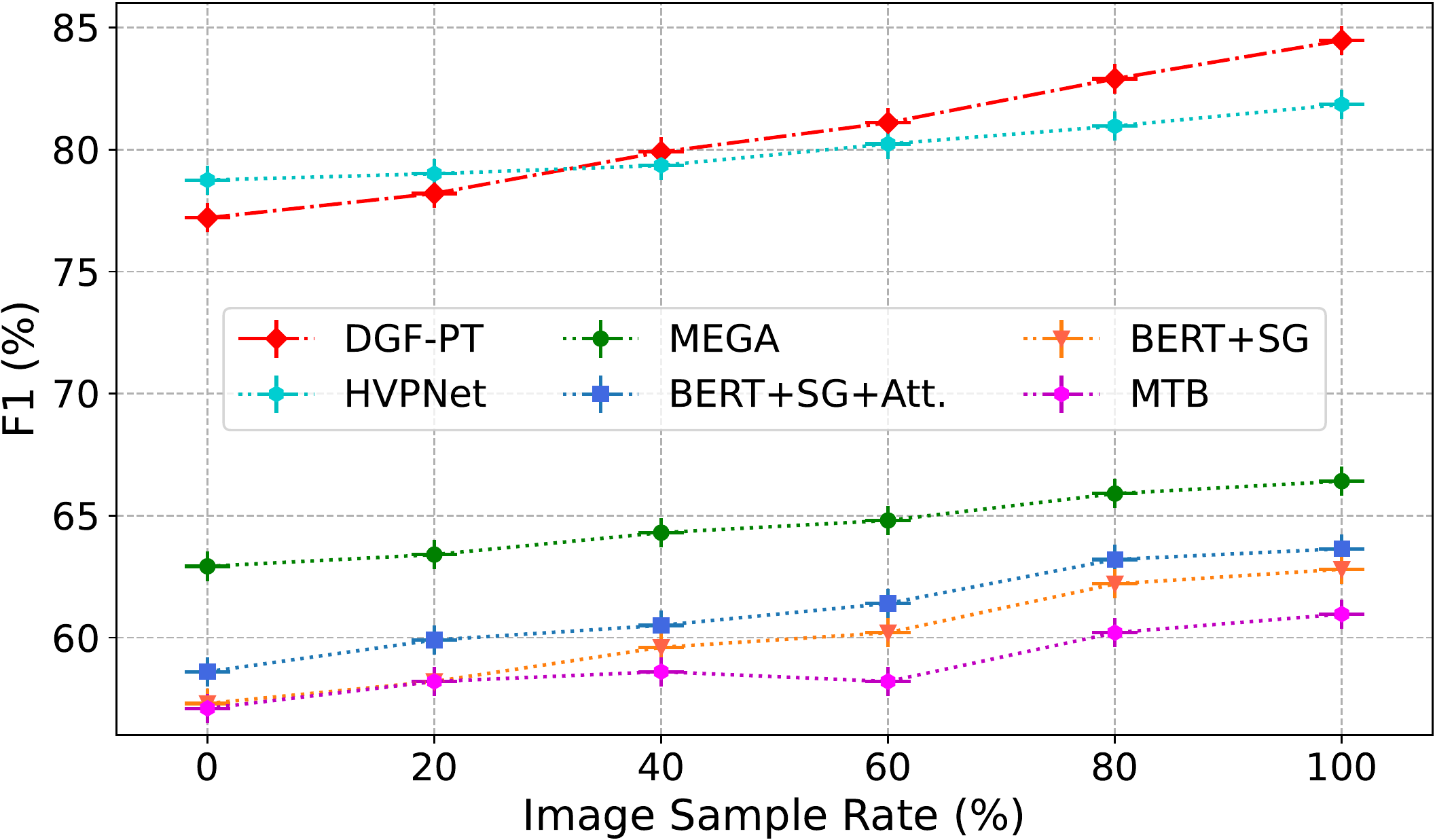}
    \caption{Different proportions of images. 
    }
    \label{graph}
\end{figure}

\subsection{Discussions for Image Information}



To further investigate the impact of images on all the compared methods, we report the results by deleting different proportions of images, as shown in Figure~\ref{graph}. From the figure we can observe that: 
1) On all metrics, the more proportion of images were introduced, the better the model performed. It demonstrates that more images provide more meaningful information for relation classification and utilize visual information more effectively.
2) Compared to other methods, our model performs best (except for HVPNet in introducing only 0\%-20\% percentage image data). 
It demonstrates that without visual information, our model is still more capable to capture intra-modal associations for relation classification.
3) Compared to the HVPNet with an object-level prefix, our model performs poorly with fewer visual data. The main reason is that the prefix is outstanding in the few-shot situation and incorporating deeper correlations of our model needs enough visual information compared to the prefix-based method. Our model performs better than HVPNet with the visual data increase. 
Observations indicate that our model incorporates visual features more effectively.


\subsection{Discussions for Sample Number}

\begin{figure}[t]
    \centering
    \includegraphics[width=\linewidth]{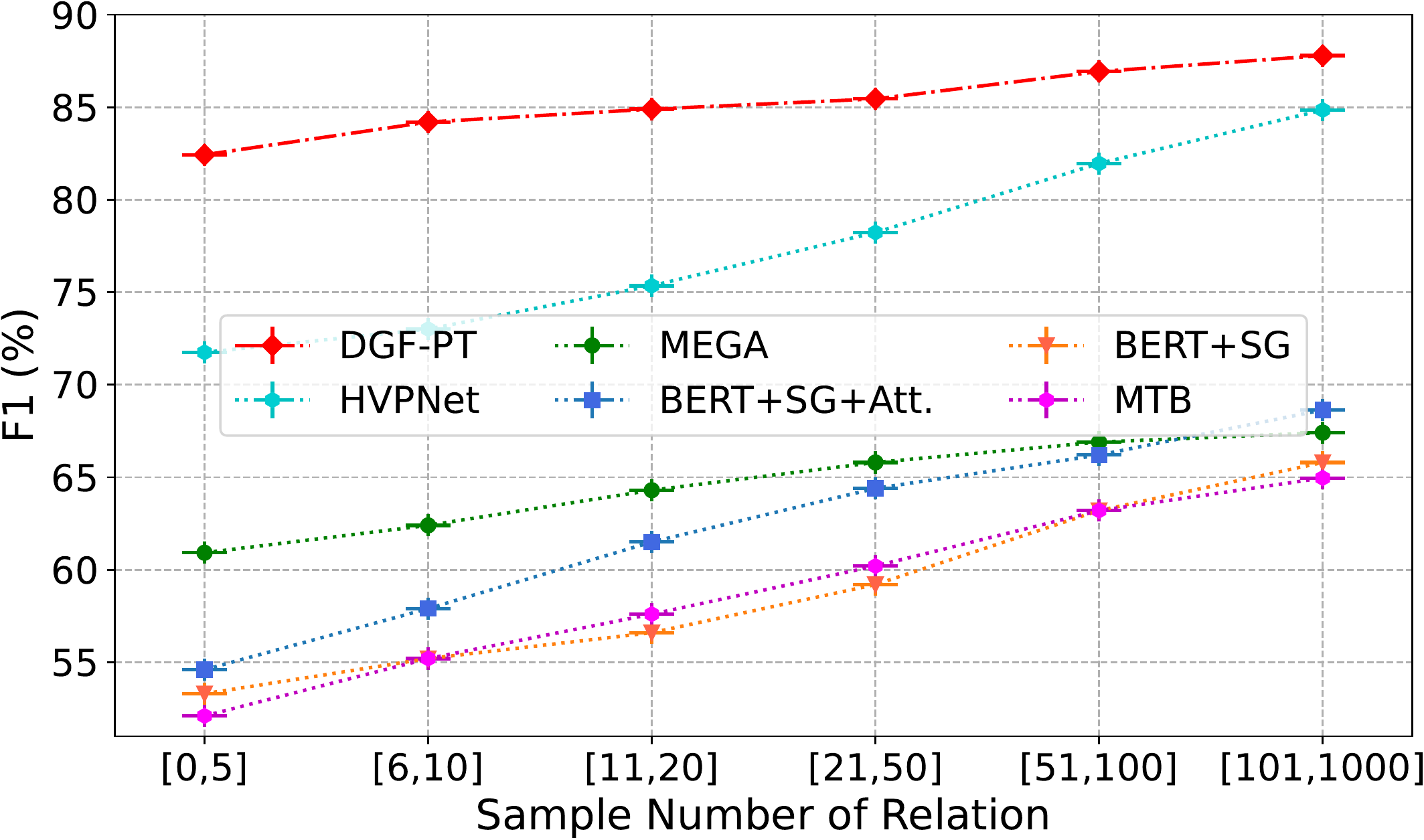}
    \caption{Impact of differences in sample number. }
    \label{relation}
\end{figure}

We investigate the impact of the sample number of different relations. 
To do so, we divide the dataset into multiple blocks based on the sample number of each relation and evaluate the performance by varying the sample number of relations in $[0,1000]$ compared with the outstanding baselines, as shown in Figure \ref{relation}.
From the figure, we can observe that: 1) The increasing of sample number  performance improvements to all methods.
The main reason is that the smaller the sample number, the more difficult it is to distinguish the relation. 2) Our model could also advance the baseline methods with the decrease in sample number, demonstrating the superiority of our method in tackling the relation with fewer samples. This phenomenon confirms the prefixes are suitable for few-shot situations. 
All the observations demonstrate that our method reduces the impact of the sample number.



\begin{figure}[t]
    \centering
    \includegraphics[width=\linewidth]{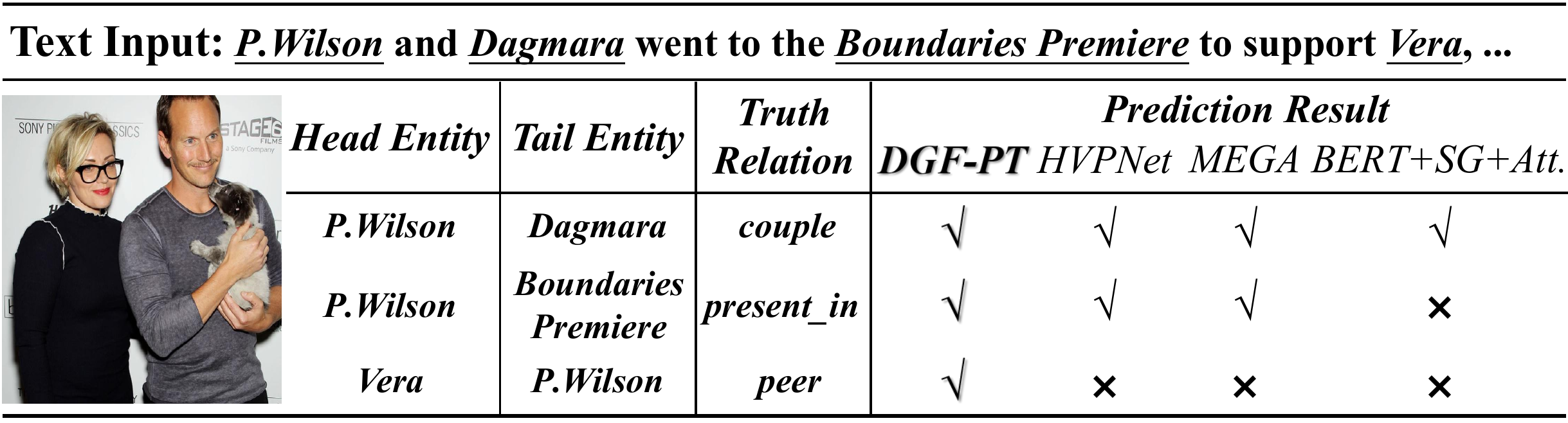}
    \caption{Case study of the same image and text with different entity pairs. \checkmark/$\mathbf{\times}$: correct/wrong prediction. 
    }
    \label{caseStudy}
\end{figure}

\subsection{Case Study}


To illustrate our model can effectively identify useful visual information, we provide an example involving various entity pairs. As shown in Fig. \ref{caseStudy}, the helpful information varies depending on the entity pair. From the figure, we can observe that: 1) Our model achieves superior performance across different entity pairs, demonstrating its ability to effectively extract useful visual information while avoiding the negative influence of unhelpful information on prediction. 2) When presented with the entity pair of \textit{Vera} and \textit{P.Wilson} that contains limited useful visual information, our model remains the best, while other baselines make incorrect predictions. 
These observations further demonstrate the effectiveness of our model in leveraging visual information while avoiding the negative influence of unhelpful information on predictions.


\section{Conclusion}

We propose DGF-PT, a novel multi-modal relation extraction framework, to capture deeper correlations among entity pair, text, and image/objects and integrate more helpful information for relation extraction. Our framework effectively integrates intra-modal and inter-modal features, distinguishes helpful visual information, and restricts candidate relations.
Extensive experiments conducted on the benchmark dataset show that our approach achieves excellent performance.

\section*{Limitations}
Our work overcomes visual noise data that limit extraction performance, incorporating multi-modal knowledge of different levels. 
Empirical experiments demonstrate that our method avoids noise data misleading the MMRE model.
However, there are still some limitations of our approach can be summarized as follows:

\begin{itemize}[leftmargin=*]
    \item Due to the limitation of the existing MMRE datasets, we only experiment on two modalities to explore the influence of image features. We will study more modalities in future work.
    \item Our method neglects the multiple relations for an input, which may not consider the multiple semantics of entities. We leave the multiple relation extraction method for future work.
\end{itemize}

\section*{Ethics Statement}
In this work, we propose a new MMRE framework that captures deeper correlations and fuses helpful visual information to benchmark our architecture with baseline architectures on the MNRE dataset.

\paragraph{Data Bias.}
Our framework is designed for multi-modal relation extraction for Twitter data. However, when applied to data with vastly different distributions or in new domains, the model's performance may be biased. The results reported in the experiment section are based on specific benchmark datasets, which may be affected by these biases. Therefore, caution should be taken when evaluating the generalizability and fairness.
\paragraph{Computing Cost/Emission.}
Our research, which entails the utilization of large language models, necessitates a significant computational burden. We recognize that this computational burden results in a negative environmental impact in terms of carbon emissions. Specifically, our work required a cumulative 425 GPU hours of computation utilizing Tesla V100 GPUs. The total emissions generated by this computational process are estimated to be 47.18 kg of $\text{CO}_2$ per run, with a total of two runs being performed.

\section*{Acknowledgment}
We thank the anonymous reviewers for their insightful comments and suggestions. 
Jianxin Li is the corresponding author.
The authors of this paper were supported by the NSFC through grant No.U20B2053, 62106059.

\bibliography{custom-simple}
\bibliographystyle{acl_natbib}

\clearpage
\appendix

\section{Oscar for Image Caption Generation}
\label{Oscar}

To generate the text description of the image for multi-modal knowledge alignment without additional pre-training on multi-modal relation extraction, we directly utilize the image captioning method, generating a natural language description of the content of an image. 
In this paper, we use Oscar (Object-Semantics Aligned Pre-training)~\cite{DBLP:conf/eccv/Li0LZHZWH0WCG20} to transfer the image into a text description for each image, which integrates multi-modal alignment information of entities from a semantic perspective.

Oscar uses object tags detected in images as anchor points to significantly facilitate alignment learning. Input samples are processed into triples involving image region features, captions, and object tags similar to the pre-training.
It randomly masks 15\% of caption tokens and uses the corresponding output representations to perform a classification to predict tokens.
Similarly to VLP \cite{DBLP:conf/aaai/ZhouPZHCG20}, the self-attention mask is constrained so that a caption token can only attend to the tokens before its position to simulate a uni-directional generation process. 
It eases the learning of semantic alignments between images and texts on the public corpus of 6.5 million text-image pairs, creating new state-of-the-art on the image caption task. Thus, we use Oscar to integrate useful images by transferring them into textual descriptions.

\section{Hyper-parameter Settings}
\label{Settings}
Our implementation is based on PyTorch\footnote{https://pytorch.org/}. All experiments were carried out on a server with one GPU (Tesla V100). For re-implementation, we report our hyper-parameter settings on the dataset in Table~\ref{tab:hyper-parameters}. 
Note that the hyper-parameter settings are tuned in the validation data by grid search with $5$ trials. The learning rate is $2e-4$, the batch size is $100$, and the dropout rate is $0.6$. We use AdamW~\cite{DBLP:conf/iclr/LoshchilovH19} to optimize the parameters. The maximum length of the text is $128$ and the objects of each image are $10$. 
For the learning rate, we adopt the method of grid search with a step size of $0.0001$. 
\begin{table}[h]
  \centering\footnotesize\setlength{\tabcolsep}{5pt}
  \begin{tabular*}{0.48 \textwidth}{l|c}
  \toprule
  Hyper-parameter & {   }{  } {   }  MNRE dataset \\
  \midrule
  word embedding dimension  {   }{   }{   }{   }  {   }{   } {   }{   }  &  {   }{  } 768   \\
  image embedding dimension     & {   }{  } 4,096  \\
  dropout rate        & {   }{  } 0.6  \\
  batch size         & {   }{  } 100  \\
  training epoch      & {   }{  } 20  \\
  maximum length of text      & {   }{  } 128  \\
 learning rate            & {   }{  } $2e-4$  \\
  threshold $\lambda_{d}$       &{   }{  }  2  \\
  threshold $\lambda_{s}$      & {   }{  } 2  \\
  threshold $\lambda_{c}$     & {   }{  } 3  \\
  threshold $\delta$ & {   }{  } 0.4  \\
  \bottomrule
  \end{tabular*}
  \caption{Hyper-parameter settings of  DGF-PT.}
  \label{tab:hyper-parameters}
\end{table}

\begin{table}[t]

\centering
\renewcommand\arraystretch{1.35}
\resizebox{\linewidth}{!}{
\begin{tabular}{l|cccc}
\toprule
 \textbf{Variants} & Acc. (\%)  &  Prec. (\%)   &  Recall (\%)  &  F1 (\%) \\
 \midrule
 \textbf{$\text{  DGF-PT (\textbf{$I_o$ $\rightarrow$ $I_i$ $\rightarrow$ $I_t$})}$}  & \textbf{85.25}  &  \textbf{84.35} & \textbf{83.83} & \textbf{84.47}   \\ \hline
  {  }{  } \textbf{$I_o$ $\rightarrow$ $I_t$ $\rightarrow$ $I_i$}   & 84.64  & 83.32  & 82.09  & 83.53  \\ 
 {  }{  } \textbf{$I_i$ $\rightarrow$ $I_o$ $\rightarrow$ $I_t$}   &  84.92 & 83.17 & 82.24 &  84.85  \\  
  {  }{  } \textbf{$I_i$ $\rightarrow$ $I_t$ $\rightarrow$ $I_o$}  & 84.85 & 83.38 & 83.70  & 84.26   \\ 
  {  }{  } \textbf{$I_t$ $\rightarrow$ $I_o$ $\rightarrow$ $I_i$ }    & 83.01  & 82.96  &82.61  & 82.73    \\
 {  }{  } \textbf{$I_t$ $\rightarrow$ $I_i$ $\rightarrow$ $I_o$}    & 82.27  & 82.04  & 81.75   & 82.29   \\ 
 
\bottomrule
\end{tabular}
}
\caption{Impact of the input order of the image $I_i$, objects $I_o$, and text $I_t$.}
\label{order}
\end{table}

\section{Discussions for Input Order}
\label{Order}
Due to utilizing a generative encoder, where the prefix, object, image, and text are input stage by stage, thus the order affects the performance of the model. As shown in Table~\ref{order}, we exploit the best input order for multi-modal relation extraction.
 From the figure, we can observe that: 1) Our model is affected by the input order of text, image, and objects. The reason we think that prompt-based autoregressive encoder is a more efficient way to integrate multi-grained information. 2) The best input order is \textbf{$I_o$ $\rightarrow$ $I_i$ $\rightarrow$ $I_t$}. Furthermore, when the text $I_t$ is input before others, the performance of our model dramatically decreases. It demonstrates that visual information is fed before textual information usually integrating more helpful extraction knowledge.



\end{document}